\pdfoutput=1

\documentclass[11pt]{article}

\usepackage[]{acl}

\usepackage{times}
\usepackage{latexsym}

\usepackage{graphicx}
\usepackage{booktabs}
\usepackage{multirow}
\usepackage{rotating}
\usepackage{amsmath}
\usepackage{amssymb}

\usepackage{bm}
\usepackage{framed}
\usepackage{caption}
\usepackage{xcolor}
\usepackage{algorithm}
\usepackage{algpseudocode}

\usepackage{soul}
\usepackage{hyperref}
\usepackage{colortbl}
\usepackage{amssymb}
\usepackage{makecell}
\usepackage{bbding}
\usepackage{enumitem}

\definecolor{esc}{HTML}{1f77b4}
\definecolor{augesc}{HTML}{d62728}
\definecolor{illogical}{HTML}{CCEBFF}
\definecolor{uninformative}{HTML}{FFE4CC}
\definecolor{non-empathetic}{HTML}{FFCCCD}

\definecolor{msgrblue}{HTML}{4889f4}
\definecolor{msgrgray}{HTML}{d3d3dc}

\newcommand{\win}[1]{{\colorbox{msgrblue}{\color{white}{\textbf{#1}}}}}
\newcommand{\lose}[1]{{\colorbox{msgrgray}{#1}}}
\newcommand{\midlose}[1]{{\colorbox{white}{#1}}}

\usepackage[utf8]{inputenc}

\usepackage{microtype}
\usepackage{inconsolata}

\title{\textsc{AugESC}: Dialogue Augmentation with Large Language Models for Emotional Support Conversation}

\author{Chujie Zheng \quad Sahand Sabour \quad Jiaxin Wen \quad Zheng Zhang \quad Minlie Huang\thanks{\ \ Corresponding author.} \\
  The CoAI Group, DCST, Institute for Artificial Intelligence, \\
  State Key Lab of Intelligent Technology and Systems, \\
  Beijing National Research Center for Information Science and Technology, \\
  Tsinghua University, Beijing 100084, China \\
  {\tt chujiezhengchn@gmail.com \quad aihuang@tsinghua.edu.cn} \\
}

\begin{document}
\maketitle
\begin{abstract}

Crowdsourced dialogue corpora are usually limited in scale and topic coverage due to the expensive cost of data curation. 
This would hinder the generalization of downstream dialogue models to open-domain topics.
In this work, we leverage large language models for dialogue augmentation in the task of emotional support conversation (ESC).
By treating dialogue augmentation as a dialogue completion task, we prompt a fine-tuned language model to complete full dialogues from available dialogue posts of various topics, which are then postprocessed based on heuristics.
Applying this approach, we construct \textsc{AugESC}, an augmented dataset for the ESC task, which largely extends the scale and topic coverage of the crowdsourced ESConv corpus.
Through comprehensive human evaluation, we demonstrate that our approach is superior to strong baselines of dialogue augmentation and that \textsc{AugESC} has comparable dialogue quality to the crowdsourced corpus.
We also conduct human interactive evaluation and prove that post-training on \textsc{AugESC} improves downstream dialogue models' generalization ability to open-domain topics.
These results suggest the utility of \textsc{AugESC} and highlight the potential of large language models in improving data-scarce dialogue generation tasks.\footnote{The project repository is available at \url{https://github.com/thu-coai/AugESC}.}

\end{abstract}

\section{Introduction}
\label{sec:introduction}

Current open-domain dialogue corpora are commonly curated through crowdsourcing to endow dialogue models with sophisticated skills \cite{esc, wow, pc}, since the desired high-quality dialogues are usually not available in public sources.
For example, the task of emotional support conversation (ESC) \cite{esc} aims to support help-seekers to reduce daily-life emotional distress.
To train downstream dialogue models, \citet{esc} crowdsourced the ESConv dataset, which contains only 1.3K dialogue sessions covering 13 topic categories.

The construction of ESConv reveals typical limitations of crowdsourcing dialogue data.
First, it is time-consuming especially when the desired dialogues should contain long multi-turn interactions (e.g., the ESConv dialogues contain about 30 utterances on average). %
Also, it usually requires laborious worker training and human screening to ensure the high dialogue quality. %
Consequently, the expensive data curation restricts the scale and topic coverage of collected dialogues, which may hinder the generalization of downstream dialogue models to open-domain topics.

\begin{figure}[t]
  \centering
  \includegraphics[width=\linewidth]{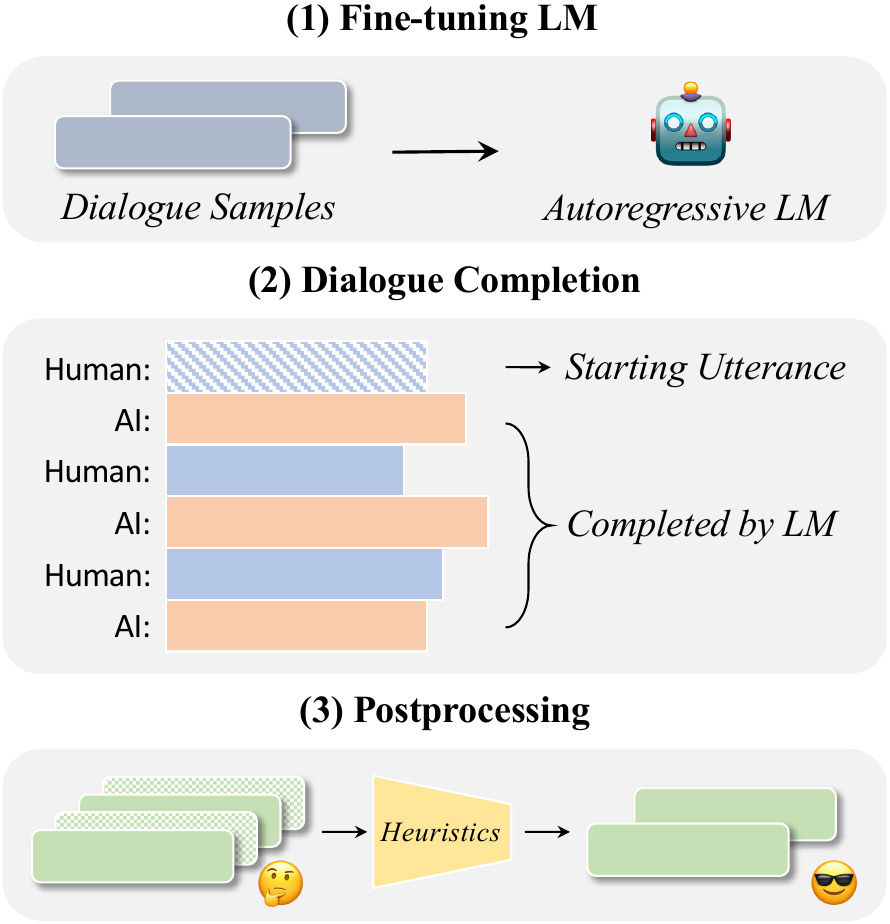}
  \caption{
    Illustration of our approach for constructing \textsc{AugESC}.
    We (1) fine-tune an autoregressive LM with dialogue samples, 
    (2) prompt the LM to complete full dialogues from collected dialogue posts, and then 
    (3) postprocess augmented dialogues based on heuristics.
  }
  \label{fig:approach}
\end{figure}

In this work, we leverage large language models for large-scale dialogue augmentation in the ESC task.
We first present a simple yet effective approach, which treats dialogue augmentation as a dialogue completion task (\S~\ref{sec:approach}), as illustrated in Figure~\ref{fig:approach}.
We fine-tune the 6B GPT-J model \cite{gptj} with ESConv samples, prompt it to complete full dialogues on various topics using the dialogue posts of EmpatheticDialogues \cite{ed}, and postprocess augmented dialogues based on heuristics.
We thus construct \textsc{AugESC}, an augmented dataset for the ESC task, which largely extends the scale (45x larger) and topic coverage of the original ESConv dataset (\S~\ref{sec:augesc}).
To demonstrate the superiority of our approach, we compare with strong baselines of dialogue augmentation (\S~\ref{sec:quality}) through human evaluation.
We show that our approach performs substantially better and that the quality of the constructed \textsc{AugESC} is comparable to the crowdsourced ESConv.
To further prove the utility of \textsc{AugESC}, we conduct human interactive evaluation  (\S~\ref{sec:utility}) and show that post-training on \textsc{AugESC} improves downstream dialogue models' generalization ability to open-domain topics.

Our contributions are summarized in four folds:
\begin{itemize}[leftmargin=*]

\item We present a simple yet effective approach for large-scale dialogue augmentation by formulating it as a dialogue completion task.

\item We release an augmented dataset \textsc{AugESC} for the ESC task, which is 45x the scale of the crowdsourced ESConv corpus and largely extends the latter's topic coverage.

\item We conduct comprehensive human evaluation, which validates the superiority of our approach and the reasonable quality of \textsc{AugESC}.

\item We conduct human interactive evaluation, which verifies \textsc{AugESC}'s utility in improving downstream dialogue models' generalization ability to open-domain topics.

\end{itemize}

\section{Related Work}
\label{sec:related}

\paragraph{Data Augmentation with Language Models}
The emergence of pre-trained language models has greatly promoted the progress of NLP technology in the past few years \cite{bert, gpt2, gpt3}, and meanwhile bring opportunities for automatic data augmentation of various NLP tasks.
For instance, \citet{schick2021generating} prompt GPT-2 \cite{gpt2} with textual instructions to generate a textual similarity dataset.
\citet{wang2021towards} leverage the 175B-parameter GPT-3 \cite{gpt3} model to generate training data for text classification and language understanding tasks.
\citet{west2021symbolic} use GPT-3 to acquire large-scale commonsense knowledge.
\citet{wanli} construct a natural language inference dataset through the collaboration with crowdworkers and GPT-3.
Different from them, our work focuses on data augmentation for open-domain dialogue generation, which is more complex and challenging due to the much longer text length, the open-ended nature, and the higher demand for dialogue quality.

The more relevant work to ours are \cite{mohapatra2020simulated, botstalk}, which both train different dialogue models to simulate the interaction between crowdworkers.
Our work differs from them in two aspects.
(1) Instead of simulated interaction, we treat dialogue augmentation as a dialogue completion task (\S~\ref{sec:approach}).
We show in \S~\ref{sec:quality} that our approach performs better in both effectiveness and efficiency.
(2) Beyond verifying the utility of augmented data in training downstream models (\S~\ref{sec:utility}), we focus more on analyzing and evaluating the quality of augmented dialogues (\S~\ref{sec:augesc} and \ref{sec:quality}).

\vspace{-1mm}
\paragraph{Emotional Support Conversation (ESC)}
ESC \cite{esc} is a dialogue generation task where the dialogue model plays the role of peer supporter and helps the help-seeker reduce daily-life emotional distress.
It usually requires various support skills to achieve effective support, such as asking questions, expressing empathy, and providing suggestions \cite{ed, comae, cem, case}.
As discussed in \cite{esc}, high-quality ESC data usually does not naturally exist and can hardly be complemented by empathetic or emotional dialogue data.
The authors thus crowdsourced the ESConv dataset through laborious worker training and quality control mechanisms.
As a result, ESConv contains only 1.3K dialogue sessions and 13 topic categories, reflecting the intractability of crowdsourcing large-scale high-quality ESC dialogues.
Our work attempts to alleviate data scarcity through automatic dialogue augmentation.

\begin{figure*}[t]
  \centering
  \includegraphics[width=\linewidth]{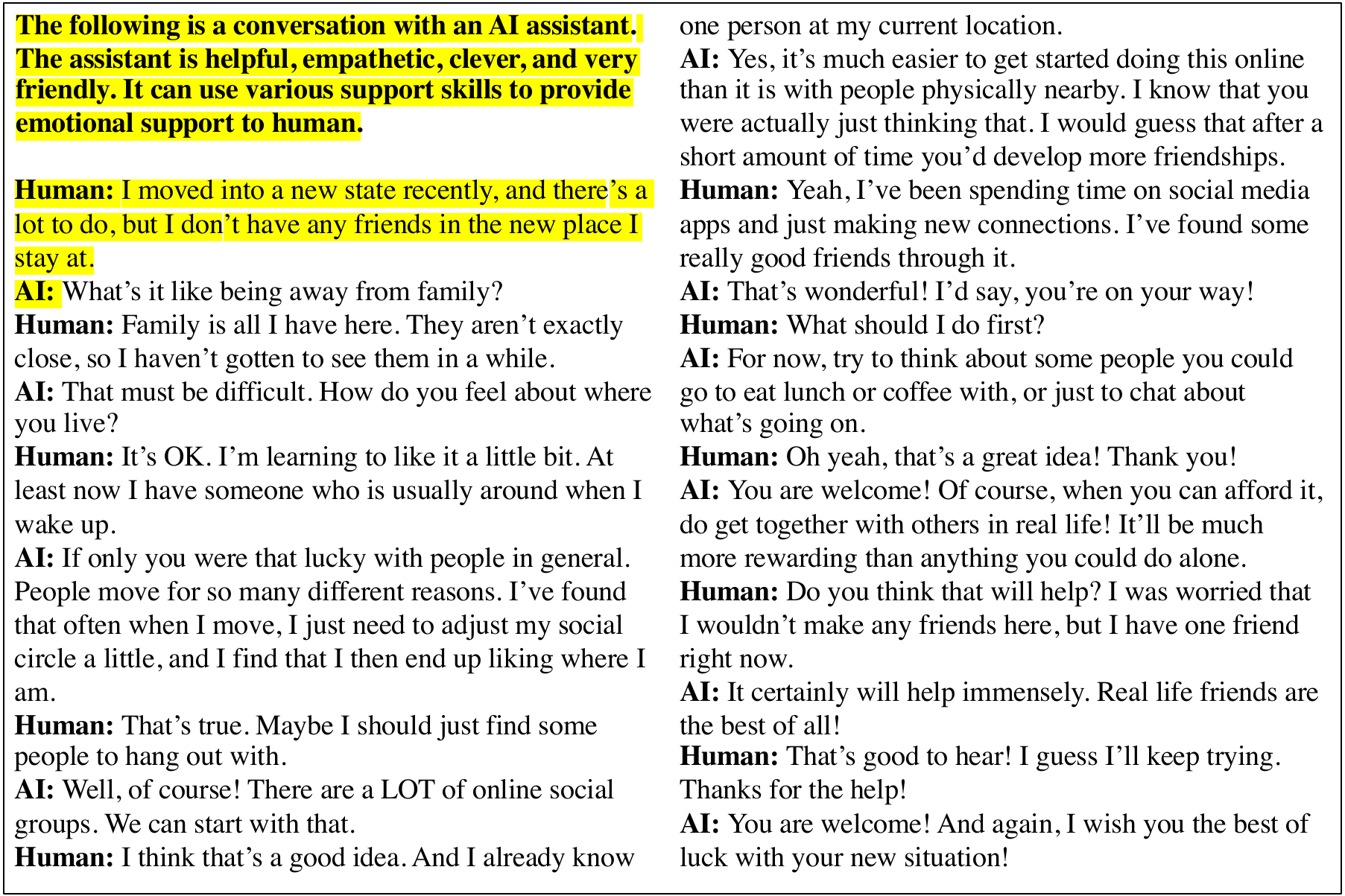}
  \caption{
    A cherry-picked example dialogue from \textsc{AugESC}.
    \colorbox{yellow}{The task description and the starting utterance} is fed into the fine-tuned language model, which then generates the subsequent dialogue.
  }
  \label{fig:data_example}
\end{figure*}

\section{Methodology}
\label{sec:approach}

We treat dialogue augmentation as a dialogue completion task, as illustrated in Figure~\ref{fig:approach} and \ref{fig:data_example}.
The dialogue augmentation procedure contains three steps:
(1) fine-tuning the language model (\S~\ref{subsec:fine-tuning}), 
(2) prompting it to complete full dialogues on various topics (\S~\ref{subsec:completion}), 
and (3) postprocessing augmented dialogues based on heuristics (\S~\ref{subsec:postprocessing}).

\subsection{Fine-tuning Language Model}
\label{subsec:fine-tuning}

\textit{As the first step of our approach, an autoregressive language model is fine-tuned with dialogue samples to acquire the ability of completing a full dialogue from the starting utterance.}

Previous work \cite{zheng2021exploring, instruction-tuning} has shown that the textual instruction facilitates the generalization of language models.
As shown in Figure~\ref{fig:data_example}, we adopts the textual instruction that contains a paragraph of task description and distinguishes the seeker and supporter with the role prompts ``Human'' and ``AI''.
During the next step of dialogue completion (\S~\ref{subsec:completion}), the language model is first fed with the task description and a starting utterance (starting with ``Human: ''), followed by the next ``AI: ''.
It then generates the subsequent dialogue until the EOS token is decoded.

In practice, we adopted GPT-J \cite{gptj}, an open-sourced autoregressive language model with 6B parameters.
We fine-tuned it for 1 epoch on 100 sampled ESConv dialogue sessions, which we found could lead to a balance between domain adaptation and the generalization to unseen dialogue topics.
See Appendix~\ref{sec:implementation} for implementation details.

\subsection{Dialogue Completion}
\label{subsec:completion}

\textit{The second step is to collect available, diverse, and informative dialogue posts as starting utterances, which are then used to prompt the language model to complete full dialogues on various topics.}

In the scope of ESC, we target those dialogue posts that describe emotional problems in daily life.
We utilized dialogue posts from EmpatheticDialogues (ED) \cite{ed}, a crowdsourced empathetic dialogue dataset widely used in academic research.
ED contains rich and diverse dialogue posts, which are assigned with emotion labels and contain detailed descriptions about the emotional states.
We used the posts with negative emotion labels and retained those with lengths between 10 and 60\footnote{We noticed that a longer post usually leads to a dialogue with longer utterances. Since the ESConv dialogues generally do not contain too long utterances, we also set an upper bound (60) for the post length to avoid large gaps with ESConv.} (with NLTK tokenization, similarly below) to ensure the proper amount of information
.
Finally, we collected 8,950 dialogue posts with the average length 19.9.
We traversed these posts for 10 epochs and obtained 89,500 raw generated texts with nucleus sampling \cite{topp} and $p=0.9$ (the default decoding algorithm adopted in our work).

\begin{table}[t]
  \centering
  \scalebox{0.85}{
    \begin{tabular}{lr}
    \toprule
    \textbf{Heuristics} & \textbf{Proportions} \\
    \midrule
    \multicolumn{2}{l}{{Augmentation Failures}} \\
    \quad\quad \textit{Non-dialogue} & 0.3\% \\
    \quad\quad \textit{Unfinished Generation} & 3.4\% \\
    \quad\quad \textit{Prompt Word Leakage} &  2.1\% \\
    \multicolumn{2}{l}{{Harmful Self-reinforcement}} \\
    \quad\quad \textit{Unbalanced \# Utterances} & 5.2\% \\
    \quad\quad \textit{Consecutive \# Utterances} & 3.5\% \\
    \multicolumn{2}{l}{{Distributional Gaps with ESConv}} \\
    \quad\quad \textit{Total \# Utterances} &  4.8\% \\
    \quad\quad \textit{Utterance Length} &  8.0\% \\
    \midrule
    {Final Retention} & 72.7\% \\
    \bottomrule
    \end{tabular}%
  }
  \caption{
  Postprocessing results for \textsc{AugESC}.
  Each middle row is the proportion of raw generated texts removed with the corresponding rules, while the bottom row is the final retention ratio.
  }
  \label{tab:postprocessing}%
\end{table}%

\subsection{Postprocessing}
\label{subsec:postprocessing}

\textit{In the final postprocessing step, we remove undesirable augmented cases based on heuristics.}

Specifically, we removed three types of undesirable cases:
(1) \textbf{Augmentation Failures}.
It includes generating \textit{Non-dialogue} contents, \textit{Unfinished Generation} where the dialogue has not been generated completely, and \textit{Prompt Word Leakage} where the utterances contain the role prompts ``Human'' or ``AI''.
(2) \textbf{Harmful Self-reinforcement}.
A language model is prone to reinforcing itself to generate new text with similar patterns to the precedent generated text \cite{topp}.
For instance, given the dialogue history where the speakers' utterance numbers are unbalanced or one speaker has said consecutive utterances, the language model is more prone to continuously generating unbalanced or consecutive utterances, which generally do not appear in natural conversations.
We thus set requirements for \textit{Unbalanced/Consecutive Utterance Number} to alleviate the harmful self-reinforced patterns, which also facilitates balanced information exchange between interlocutors.
(3) \textbf{Distributional Gaps with ESConv}.
We also set requirements for the \textit{Total Utterance Number} and \textit{Utterance Length} to (a) avoid large distributional gaps with ESConv and (b) encourage in-depth discussion with enough conversation turns.
See Appendix~\ref{sec:rules} for details of the filtering rules.

Table~\ref{tab:postprocessing} shows the postprocessing results.
After postprocessing the 89,500 raw texts, we obtained 65K augmented dialogues (72.7\%).
Benefiting from model fine-tuning, our approach shows a good controllability of generating valid dialogues (only 0.3\% non-dialogue, 3.4\% unfinished generation, and 2.1\% prompt word leakage).
The phenomenon of self-reinforcement also unavoidably occurs in the generation of GPT-J (5.2\%/3.5\% unbalanced/consecutive utterance number), which suggests the necessity of restricting the utterance numbers during postprocessing.

Considering that heuristic-based postprocessing may not guarantee the perfect quality of augmented dialogues, we also conducted comprehensive human evaluation to assess the dialogue quality (\S~\ref{sec:quality}).
The results show that the currently obtained \textsc{AugESC} has been of reasonable quality.
On the other hand, there are still no reliable automatic methods for open-domain dialogue evaluation \cite{not-evaluate, deriu2021survey}, and even human evaluation is an open problem \cite{human-evaluation}.
We leave further quality refinement methods of dialogue augmentation for future work.

\begin{table}[t] 
  \centering
  \scalebox{0.85}{
    \begin{tabular}{llcc}
    \toprule
    \multicolumn{2}{c}{} & \textbf{ESConv} & \textbf{\textsc{AugESC}} \\
    \midrule
    \multicolumn{2}{l}{\# Sessions} & 1.3K & \textbf{65K} \\
    \multicolumn{2}{l}{Average Session Length} & 543.6  & \textbf{496.4} \\
    \multicolumn{2}{l}{\# Utterances} & 38K & \textbf{1,738K} \\
    \multicolumn{2}{l}{Average \# Utterances} & 28.9  & \textbf{26.7} \\
    \multicolumn{2}{l}{Average Utterance Length} & 18.8  & \textbf{18.7} \\
    \midrule
    \multirow{3}[0]{*}{Seeker} & \# Utterances & 20K & \textbf{867K} \\
       & Avg \# Uttr & 15.4  & \textbf{13.3} \\
       & Avg Uttr Len & 16.8  & \textbf{17.4} \\
    \midrule
    \multirow{3}[0]{*}{Supporter} & \# Utterances & 18K & \textbf{872K} \\
       & Avg \# Uttr & 13.6  & \textbf{13.4} \\
       & Avg Uttr Len & 21.0  & \textbf{19.8} \\
    \bottomrule
    \end{tabular}%
  }
  \caption{
  Statistics of \textsc{AugESC} compared with ESConv.
  For ESConv, we removed utterances from supporters at the beginning of dialogues because these utterances are usually uninformative greetings.
  }
  \label{tab:statistics}%
\end{table}

\section{Data Analysis}
\label{sec:augesc}

\subsection{Statistics}
\label{subsec:statistics}

The statistics of our constructed \textsc{AugESC} dataset are shown in Table~\ref{tab:statistics}.
An example dialogue from \textsc{AugESC} is shown in Figure~\ref{fig:data_example}.
\textsc{AugESC} contains 65K dialogue sessions and 1,738K utterances, roughly 50/45x the scale of ESConv.
The \textsc{AugESC} dialogues generally have a close utterance number and length to ESConv due to the heuristics for controlling their distributional gaps (Table~\ref{tab:postprocessing}, Total Utterance Number and Utterance Length).
We observe that in the ESConv dialogues, the utterance number of seeker is usually larger than supporter (15.4 vs. 13.6), while their numbers are closer in \textsc{AugESC} (13.3 vs. 13.4).
This is because the augmentation process of \textsc{AugESC} has to trade off the self-reinforcement phenomenon, as discussed in \S~\ref{subsec:postprocessing} (the heuristics of Unbalanced/Consecutive Utterance Number).

\begin{table}[t]
  \centering
  \scalebox{0.85}{
    \begin{tabular}{p{22em}}
    \toprule
    \multicolumn{1}{c}{\textbf{ESConv}} \\
    \midrule
    pandemic (5.2), covid (5.0), depression (3.4), support (3.2), christmas (3.1), 
    job (2.6), anxiety (2.6), online (2.6), vaccine (2.4), zoom (2.2), 
    holidays (2.2), correct (2.1), feeling (2.1), helpful (2.1), stress (2.0), 
    virus (2.0), hard (2.0), breakup (2.0), mturk (1.9), merry (1.9), 
    quit (1.9), virtual (1.8), unemployment (1.8), struggling (1.8), resume (1.8), 
    youtube (1.7), honestly (1.7), moment (1.7), daily (1.6), survey (1.6) \\
    \midrule[0.3mm]
    \multicolumn{1}{c}{\textbf{\textsc{AugESC}}} \\
    \midrule
    car (4.9), sounds (4.0), dog (3.2), guess (3.2), house (2.9), 
    police (2.8), money (2.6), parents (2.6), hope (2.5), brother (2.5), 
    idea (2.4), buy (2.4), neighbors (2.4), insurance (2.3), afraid (2.2), 
    mom (2.1), luck (2.1), driving (2.1), agree (2.0), told (2.0), 
    husband (2.0), excited (2.0), Figure~(2.0), nice (1.9), upset (1.9), 
    cat (1.9), sense (1.9), scared (1.9), vet (1.8), stole (1.8) \\
    \bottomrule
    \end{tabular}%
    }
  \caption{
  Top 30 salient topic features associated with ESConv and \textsc{AugESC}.
  The rounded $z$-scored log odds ratios are marked in the parentheses, where values greater than 2 indicate significant ($>2$ std) association.
  }
  \label{tab:topic}%
\end{table}%

\subsection{Topic Analysis}
\label{subsec:topic}

To analyze the topic features, we extracted the lexical correlates of ESConv and \textsc{AugESC}.
We calculated the log odds ratio, informative Dirichlet prior \cite{monroe2008fightin} of all words for each dataset contrasting to the other.
Statistics are based on the whole 1.3K ESConv sessions or the randomly sampled 1.3K \textsc{AugESC} sessions for fair comparison.
As shown in Table~\ref{tab:topic}, dialogue topics in ESConv are closely related to its period of data curation (2020 to 2021).
For instance, the topics like ``\textit{pandemic, covid, vaccine, virus}'' are directly relevant to COVID-19, while those like ``\textit{online, zoom, virtual}'' imply the influence of COVID-19.
It indicates that the ESConv dialogues may revolve around the emotional problems under the background of or caused by COVID-19, as demonstrated by the topics ``\textit{depression, job, anxiety, breakup, unemployment}''.
As a result, ESConv may fail to cover topics about general daily life (besides, the ESConv dialogues also leak the information of crowdsourcing tasks, such as ``\textit{mturk, quit, survey}'').
By contrast, \textsc{AugESC} covers a broader range of daily-life dialogue topics, such as ``\textit{car, dog, house, police, money}'' and many others in Table~\ref{tab:topic}, benefiting from the diverse dialogue posts from ED (\S \ref{subsec:completion}).
We thus suggest that \textsc{AugESC} largely complements the topic coverage of ESConv, which can facilitate the generalization of downstream dialogue models to open-domain topics.

\subsection{Diversity Analysis}
\label{subsec:diversity}

\begin{figure}[t]
  \centering
  \includegraphics[width=\linewidth]{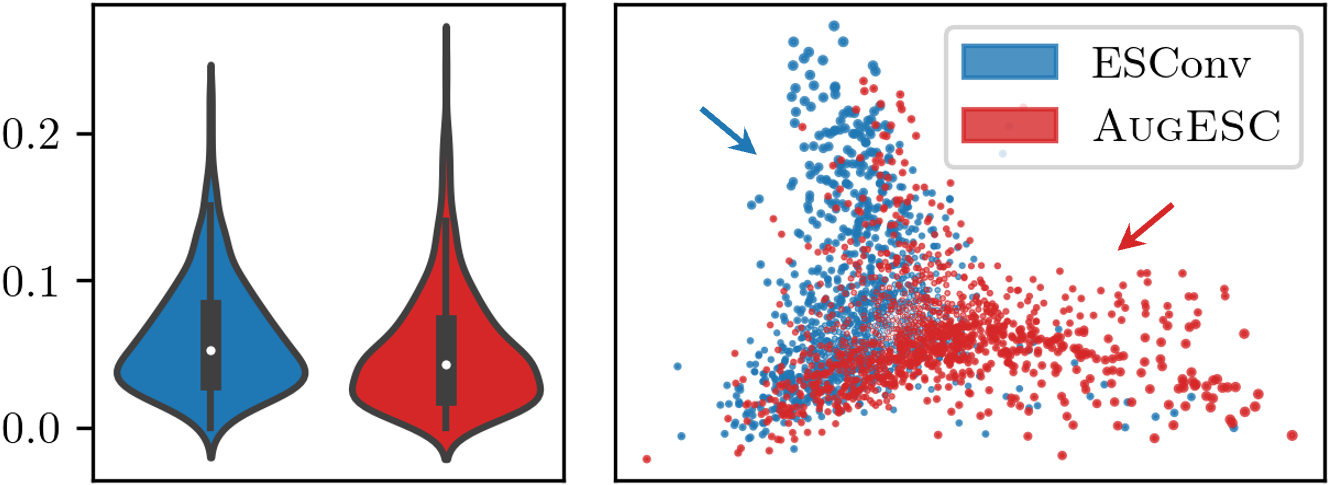}
  \caption{
  \textbf{Left}: Statistics of inter-dialogue similarity, calculated with the TF-IDF vectors of dialogues.
  \textbf{Right}: Visualization of the TF-IDF vectors of dialogues by applying 2-dimensional Principal Component Analysis (PCA).
  We mark the regions that are specially correlated with the two datasets respectively.
  }
  \label{fig:tfidf}
\end{figure}

To analyze the diversity of augmented dialogues, we extracted the TF-IDF vector of each dialogue in ESConv and \textsc{AugESC} using the Sklearn library \cite{pedregosa2011scikit}.
We computed and counted the TF-IDF similarity between any two dialogues.
As shown in Figure~\ref{fig:tfidf} (left), \textcolor{augesc}{\textsc{AugESC}} has close inter-dialogue diversity to \textcolor{esc}{ESConv}.
It suggests that different dialogues in \textsc{AugESC} have little overlap with each other, which thus can provide diverse training samples for downstream dialogue models.

We also visualized the TF-IDF vectors of dialogues by applying 2-dimensional Principal Component Analysis (PCA).
As shown in Figure~\ref{fig:tfidf} (right), the regions that are specially correlated with the two datasets are salient, as marked explicitly with arrows.
It suggests that \textsc{AugESC} has a different range of dialogue contents from ESConv and their combination can lead to a broader range.

\begin{table*}[t]
  \centering
  \scalebox{0.85}{
    \begin{tabular}{lcccccc}
    \toprule
    & \small \textbf{Informativeness} & \small \textbf{Understanding} & \small \textbf{Helpfulness} & \small \textbf{Consistency} & \small \textbf{Coherence} & \small \textbf{Unsafety ↓} \\
    \midrule
    $\kappa$ &  0.42  &  0.35   &  0.37  &  0.35  &  0.40  & 0.39  \\
    \midrule
    \textit{Crowdsourced} &   \textit{2.52}  &  \textit{2.42}   &  \textit{2.23}  &  \textit{2.56}  & \textit{2.42}    & \textit{0.13} \\
    Simulated Chat (BlenderBot) &  ~~\lose{1.86}* & ~~\lose{1.90}* & ~~\lose{1.49}* & ~~\lose{2.12}* & ~~\lose{1.90}* & \win{0.03} \\
    Simulated Chat (GPT-J) & 2.29 & 2.28 & 2.02 & ~~\midlose{2.25}* & ~~\midlose{2.16}* & 0.14 \\
    Our Approach w/o FT (GPT-3) &  ~~\midlose{2.23}*  & ~~\lose{2.07}*  & ~~\lose{1.62}*  & ~~\lose{2.11}*  & ~~\lose{1.96}*   & 0.16 \\
    \textbf{Our Approach} &  \win{2.41}  &  \win{2.37}  &   \win{2.12}  &  ~~\win{2.34}*  &  ~~\win{2.19}*    & 0.14 \\
    \bottomrule
    \end{tabular}%
  }
  \caption{
  Human evaluation results for dialogue quality.
  The scores (from 0 to 3) are averaged over all the samples rated by three annotators.
  $\kappa$ denotes Fleiss' Kappa \cite{fleiss-kappa}, indicating fair or moderate inter-annotator agreement ($0.2 < \kappa < 0.6$).
  The highest scores of \textit{augmented dialogues} are \win{highlighted} and the significantly worse ones are marked with \lose{gray background} (Student's t-test, $p$-value $<0.01$).
  * denotes significant gaps with the \textit{crowdsourced ESConv}.
  We did not conduct significance test with Unsafety due to only a few non-zero scores.
  }
  \label{tab:quality}%
\end{table*}%

\section{Evaluation for Dialogue Quality}
\label{sec:quality}

The quality of dialogue data is critical to training downstream dialogue models \cite{eva, eva2}.
To validate the quality of \textsc{AugESC}, we conduct comprehensive human evaluation and compare our approach with strong baselines of dialogue augmentation.

\subsection{Compared Methods}
\label{subsec:baseline}

\paragraph{Simulated Chat}
This baseline comes from \cite{mohapatra2020simulated, botstalk}, which simulates the crowdworkers' interaction with separately trained models.
Following \cite{mohapatra2020simulated}, we fine-tuned two models with the seekers' and supporters' utterances in ESConv separately.
We used the dialogue posts in \S~\ref{subsec:completion} as the first seekers' utterances, and then the two models took turns to reply to simulate interaction.
We set the maximum utterance number to 40 and terminated the simulated conversation if the latest utterance contained the word ``bye''.
We implemented it with two different base models.
\textbf{Simulated Chat (BlenderBot)} uses the 1.4B BlenderBot model \cite{blender}, which is the state-of-the-art open-sourced dialogue model.
\textbf{Simulated Chat (GPT-J)} uses the same 6B GPT-J model as in our approach.

\vspace{-1mm}
\paragraph{Our Approach w/o FT}
This baseline denotes directly prompting the language model to complete dialogues without fine-tuning, which is an ablated variant of our approach and can help us understand the influence of fine-tuning in \S~\ref{subsec:fine-tuning}.
However, we found that GPT-J cannot work well without fine-tuning. 
We thus implemented this ablated variant with the 175B \textbf{GPT-3} \texttt{davinci} model \cite{gpt3} through the OpenAI API.
We applied the same postprocessing as in \S~\ref{subsec:postprocessing}. See Appendix~\ref{sec:gpt3} for details.

\vspace{-1mm}
\paragraph{Crowdsourced}
We meanwhile evaluate the quality of the ESConv dialogues, which are written by crowdworkers and contain natural human-human interactions.
We expect this baseline to serve as an anchor point of dialogue quality evaluation.

\subsection{Evaluation Setups}
\label{subsec:quality-setup}

We refer to \cite{esc} to design the evaluation protocol.
When crowdsourcing the ESConv dataset, \citet{esc} asked the crowdworkers to complete a survey to rate their performance during conversation.
The survey results were used to build up the data screening criteria of ESConv.
Inspired by this survey, we design the following metrics for assessing the quality of augmented dialogues.
\textbf{Informativeness} measures how much detailedly the help-seeker describes his/her emotional problem.
\textbf{Understanding} measures how much the supporter understands the help-seeker's experience and feelings.
\textbf{Helpfulness} measures how much the supporter succeeds in helping the seeker reduce emotional distress and feel better.

Besides, we also assess the general dialogue quality.
\textbf{Consistency} measures whether the behaviors of the speakers are consistent with their roles, and whether the behavior of the same speaker is not self-contradictory.
\textbf{Coherence} measures whether the conversation is on-topic and in-depth and whether the topic transition is natural.
\textbf{Unsafety} measures whether the conversation contains unsafe contents, such as toxic language, sensitive topics, or immoral suggestions.
See Appendix~\ref{sec:guideline} for the detailed evaluation guideline.

All the metrics are rated with the four-level Likert scale ranging from 0 to 3 (higher is better except Unsafety).
We recruited 60 college students as annotators.
We randomly sampled 60 dialogue sessions for each method.
Each session was rated by three different annotators.

\subsection{Results}
\label{subsec:quality-result}

As shown in Table~\ref{tab:quality}, our approach produces augmented dialogues with the highest quality among all the methods.
(1) \textbf{Compared to Simulated Chat (BlenderBot)}, our approach demonstrates the better potential of general language models for dialogue augmentation than specifically pre-trained dialogue models (our advantages are reflected in almost all the metrics).
It is because general language models store more general knowledge and thus can better generalize to various dialogue posts for dialogue augmentation.
(2) \textbf{Compared to Simulated Chat (GPT-J)}, our advantage is not substantial due to the same base model.
However, since our approach performs one-time generation of the whole dialogue, it is superior in efficiency to Simulated Chat, which takes turns to generate the two speakers' utterances with two models (as a reference, time cost: 60 seconds vs. 80 seconds per session, GPU memory: 1 model vs. 2 models).
Interestingly, our approach slightly outperforms Simulated Chat, which indicates that fine-tuning one language model with whole dialogues may be better than fine-tuning two models with two speakers' utterances separately (the former can be viewed as the parameter-shared version of the latter).
(3) \textbf{Compared to Our Approach w/o FT (GPT-3)}, our approach is remarkably better due to model fine-tuning, which enables better adaptation to our interested ESC task, as reflected in the metrics Understanding and Helpfulness.
(4) \textbf{Compared to the crowdsourced ESConv}, our approach produces augmented dialogues with comparable evaluation scores, which confirms the reasonable quality of \textsc{AugESC}.
Nevertheless, there is still a gap in Consistency and Coherence, which is a long-standing problem in open-ended text generation \cite{scarecrow} and deserves further study in the more general NLG research.

See Appendix~\ref{subsec:quality} for further discussion about the limitations of \textsc{AugESC}'s quality.
Also see Appendix~\ref{sec:baseline-example} for additional augmented dialogue examples of Simulated Chat (BlenderBot) and Our Approach w/o FT, respectively.

\begin{table}[t]
  \centering
  \scalebox{0.85}{
    \begin{tabular}{lcc}
    \toprule
    \textbf{Attributes} & \textbf{ESConv} & \textbf{\textsc{AugESC}} \\
    \midrule
    Toxicity & 0.0613  & \win{0.0597}  \\
    Severe Toxicity & 0.0427  & \win{0.0418}  \\
    Identify Attack & {0.0531}  & \win{0.0493}  \\
    Insult & \win{0.0786}  & {0.0823}  \\
    Profanity & {0.0539}  & \win{0.0516}  \\
    Threat & 0.1155  & \win{0.1146}  \\
    \bottomrule
    \end{tabular}%
  }
  \caption{
    Results of toxicity assessment using Perspective API.
    Lower scores are better and are \win{highlighted}.
  }
  \label{tab:toxicity}%
\end{table}%

\subsection{Toxicity Assessment}
\label{subsec:toxicity}

We manually inspected the few cases with non-zero unsafety scores (all are 1-point) in augmented and crowdsourced dialogues and found that they are all about potentially sensitive topics (e.g., talking about alcohol experiences).
We conjecture that the unsafety of Simulated Chat (BlenderBot) is the lowest (even lower than the Crowdsourced ESConv) because the BlenderBot model has been processed by safety mechanisms \cite{xu2020recipes} and usually avoids discussion about sensitive topics.
However, this also sacrifices the quality of augmented dialogues, as reflected in the lowest quality of Simulated Chat (BlenderBot) in Table~\ref{tab:quality}.

Language toxicity has been an essential consideration in the NLP research \cite{realtoxicityprompts, dinan2021anticipating, click}.
As an additional evaluation for the potential toxicity in augmented dialogues, we assessed ESConv and \textsc{AugESC} using Perspective API, a widely used toxicity detection API in online discussions.
All the utterances were assessed by Perspective API's toxicity scores (between 0 and 1, lower is safer) of six production attributes.
For each attribute, we reported the score averaged over all the utterances.
From Table~\ref{tab:toxicity}, ESConv and \textsc{AugESC} both show little toxicity (very low scores) and \textsc{AugESC} is even lower, while the dialogue quality of \textsc{AugESC} still slightly underperforms ESConv, as judged by human annotators (Table~\ref{tab:quality}).
We conjecture that there may be a trade-off between reducing such ``toxicity'' and improving dialogue quality.
For instance, the supporter cannot help the help-seeker reduce emotional distress without discussing the seeker's emotional problem in depth.
On the other hand, toxicity in dialogues is subtle due to its sensitiveness to the dialogue context \cite{toxicity-context, diasafety}, and its identification is still under exploration.
Given that the toxicity detectors like Perspective API may introduce new biases \cite{hosseini2017deceiving, sap2019risk}, we did not apply additional toxicity postprocessing to \textsc{AugESC} in the current work.
We leave the further investigation of the toxicity problem in augmented dialogues as future work.

\section{Evaluation for Data Utility}
\label{sec:utility}

Recall that dialogue augmentation aims to improve smaller downstream dialogue models, which is a realistic and practical setting since the deployment of large language models is expensive.
To verify the utility of \textsc{AugESC}, we conduct \textbf{human interactive evaluation} to explore how much \textsc{AugESC} can improve the generalization of downstream dialogue models to open-domain topics.

\subsection{Compared Models}

We compared two 1.4B BlenderBot models: one is fine-tuned \textbf{only on ESConv} (1,100 dialogues, 2 epochs), while the other is further \textbf{post-trained on \textsc{AugESC}} (1 epoch).
Note that we used \textsc{AugESC} for model post-training since we observed that (1) post-training on \textsc{AugESC} facilitates better generalization to open-domain topics, and (2) first-training on ESConv provides a good initialization point due to the better consistency and coherence of crowdsourced dialogues (Table~\ref{tab:quality}), which is critical to the multi-turn interaction capability.

\begin{table}[t]
  \centering
  \scalebox{0.85}{
    \begin{tabular}{ccc}
    \toprule
    Trained on \textsc{AugESC}? & Yes & No \\
    \midrule
    \textbf{Fluency} & \win{47} & \lose{13}  \\
    \textbf{Identification} & \win{68} & \lose{22} \\
    \textbf{Comforting} & \win{55} & \lose{22} \\
    \textbf{Suggestion} & \win{58} & \lose{15} \\
    \cmidrule{1-3}
    \textbf{Overall} & \win{58} & \lose{28} \\
    \bottomrule
    \end{tabular}%
  }
  \caption{
  Results of human interactive evaluation under the open-domain setting (winning ratios shown).
  All the gaps are statistically significant (sign test, $p$-value $<0.05/0.01$ for ``Overall'' / other metrics).
  }
  \label{tab:interactive}%
\end{table}%

\subsection{Evaluation Setups}

Following \cite{esc}, we conducted pairwise human interactive evaluation.
We recruited the same participants as in \S~\ref{sec:quality}.
Each participant was asked to talk about the same emotional problem with the two bots, which accepted the same first utterances.
Each conversation lasted at least 8 turns (8 utterances from participants and 8 from bots), after which the participant could either continue or end it.
It is worth noting that we adopted the \textbf{open-domain setting}, that is, the participants were allowed to talk about any topics they wanted without restrictions.
It is distinct from the setting in \cite{esc} where the participants were only allowed to talk about limited topic categories (i.e., in-domain topics).

After the conversations, the participants were asked to compare the two bots based on the following aspects, which follow the evaluation protocol of \cite{esc}.
\textbf{Fluency}: which bot's responses were more fluent and understandable?
\textbf{Identification}: which bot explored your situation more in-depth and was more helpful in identifying your problem?
\textbf{Comforting}: which bot was more skillful in comforting you?
\textbf{Suggestion}: which bot gave you more helpful suggestions for your problems?
\textbf{Overall}: generally, which bot's emotional support do you prefer?
We collected 60 pairs of interactive conversations (each participant contributed one).

\subsection{Results}

As shown in Table~\ref{tab:interactive}, \textsc{AugESC} significantly improves the dialogue model's performance in all aspects.
It strongly proves the utility of \textsc{AugESC} in enhancing the generalization of downstream dialogue models to open-domain topics.
We argue that the results are \textit{non-trivial}.
The BlenderBot model has been very powerful in open-domain dialogue \cite{blender}. 
Given that the crowdsourced ESConv dataset has greatly improved BlenderBot's ability of emotional support \cite{esc}, any further substantial improvement on top of this is by no means easy.

\begin{table}[t]
  \centering
  \scalebox{0.85}{
    \begin{tabular}{ccccccc}
    \toprule
    \textsc{AugESC}? & \textbf{PPL} & \textbf{B-2} & \textbf{B-4} & \textbf{R-L} & \textbf{D-2} & \textbf{D-3} \\
    \midrule
    No & 11.2 & 7.8 & 2.4 & 16.9 & 23.8 & 48.0 \\
    Yes & 11.5 & 7.7 & 2.4 & 16.7 & 24.3 & 49.4 \\
    \bottomrule
    \end{tabular}%
  }
  \caption{
  Results of automatic evaluation on the in-domain ESConv test set. 
  Metrics include perplexity (PPL), BLEU-2/4 \cite{bleu}, ROUGE-L \cite{rouge} and Distinct-2/3 \cite{distinct}.
  }
  \label{tab:auto}%
\end{table}%

\subsection{In-domain Evaluation}

We conducted the additional automatic evaluation on the in-domain ESConv test set (200 held-out dialogue sessions), which aims to verify whether post-training on \textsc{AugESC} sacrifices the in-domain performance.
As shown in Table~\ref{tab:auto}, post-training on \textsc{AugESC} little influences the in-domain performance (the gaps are marginal), suggesting that improving the open-domain generalization ability is compatible with maintaining the underlying dialogue capability.

\section{Conclusion}

In this work, we present a simple yet effective approach for dialogue augmentation, which is formulated as a dialogue completion task.
Using this approach, we release an augmented dataset \textsc{AugESC} for the task of emotional support conversation (ESC), which largely extends the scale and topic coverage of the crowdsourced ESConv corpus.
Through comprehensive empirical evaluation, we show that:
(1) our approach produces augmented dialogues with higher quality than strong baselines of dialogue augmentation,
(2) \textsc{AugESC} has comparable dialogue quality to the crowdsourced ESConv corpus, and
(3) post-training on \textsc{AugESC} notably improves the generalization capability of downstream dialogue models to open-domain topics.
Our work demonstrates the prowess and utility of large language models in improving data-scarce tasks, especially complex open-domain dialogue tasks. 
It may inspire more work regarding training data augmentation with large language models.
Future work can explore automatic methods for further quality refinement of dialogue augmentation.

\section*{Ethical Considerations}
\label{sec:ethic}

The EmpatheticDialogues \cite{ed} dataset for dialogue post collection, the GPT-J model \cite{gptj}, and the BlenderBot model \cite{blender} are all widely used in academic research, can be accessed from HuggingFace Hub or official websites, and are all in the English language as well as \textsc{AugESC}.
Using the above public resources, the construction of \textsc{AugESC} does not involve human participants and thus does not collect any personal identifying information.

We raise attention that \textsc{AugESC} may possibly contain toxic or biased contents, which cannot be fully assessed in either automatic or human evaluation (\S~\ref{sec:quality}).
Future access to \textsc{AugESC} should be only for research usage and should \textbf{NOT} be used for real-deployed systems, commercial purposes, or any other usage than academic research.
Anyone using \textsc{AugESC} in the research should be aware of its limitations and should acknowledge and/or try to mitigate them to the extent possible.

Our work strictly follows the task definition and evaluation protocols (\S~\ref{sec:quality} and \ref{sec:utility}) of the original ESC paper \cite{esc}, where the support is provided through social interactions (e.g., between peers or friends) rather than professional counseling.
As mentioned in \cite{esc}, further efforts are still needed to probe the ethical extent to which dialogue models can or should provide support.
These protocols should also not be used directly in fields other than the ESC task (i.e., peer emotional support in daily life) that require the guidance of professional researchers, such as psychological counseling.

We also ethically conducted the human evaluation.
We transparently communicated with the participants of our study intent and explicitly informed them of the disclaimers before they participated.
We paid the participants at the hourly wage above \$10/hour, going well beyond the local labor compensation standard.
We acknowledge that the results of human evaluation could be affected by the participants' demographic and geographic characteristics.
This work has obtained study approval from the Institutional Review Board (IRB).

\section*{Acknowledgements}

This work was supported by the National Science Foundation for Distinguished Young Scholars (with No. 62125604). This work was also supported by the Guoqiang Institute of Tsinghua University, with Grant No. 2020GQG0005, Tsinghua Precision Medicine Foundation, and the NSFC project (with No. 62206150).

\bibliography{custom}
\bibliographystyle{acl_natbib}

\clearpage
\appendix

\section{Limitations}
\label{sec:limitation}

\subsection{Dialogue Quality of \textsc{AugESC}}
\label{subsec:quality}

Through our manual inspection, we found that the \textit{inconsistency} issue mainly occurs in the seeker-provided information.
For instance, the seeker first expresses ``\textit{sadness about the loss of the dog that he/she has raised for 14 years}''.
When the supporter asks ``\textit{the age of the dog}'', the seeker answers ``\textit{13}'', which is obviously contradictory to the aforementioned ``\textit{14 years}''.
While the inconsistency can be easily detected based on human commonsense, models are prone to make such mistakes \cite{dnli, decode, cdconv}.

Another issue we noticed is the \textit{improper topic transition}.
That is, after several turns of conversation, the supporter sometimes discusses topics other than the seeker's emotional problem.
We conjecture that the root cause is the seeker's inability to proactively provide personalized, in-depth, and detailed information about the emotional problem.
In this case, the conversation is only driven by the suggestions offered or the questions raised by the supporter, which may thus induce improper topic transition and make the conversation less coherent.

\subsection{Generalization to Other Tasks or Models}
\label{subsec:generalization}

We only experimented with the ESC task as our work focuses more on the quality analysis (\S~\ref{sec:augesc} and \ref{sec:quality}) and utility evaluation (\S~\ref{sec:utility}) of augmented dialogues.
The motivation we study in the ESC task is in two folds:
(1) The construction of the ESConv dataset reveals the typical limitations of crowdsourcing dialogue corpora (\S~\ref{sec:introduction}).
(2) \cite{esc} provides detailed data screening criteria, enabling us to design a reasonable and convincing protocol for dialogue quality evaluation (\S~\ref{sec:quality}).
However, it is intuitive to generalize our approach to other dialogue generation tasks.
For instance, in knowledge-grounded dialogue \cite{wow, kdconv, diffks}, we can additionally utilize knowledge bases for dialogue augmentation.
We leave the broader applications of our dialogue augmentation approach to future work.

While we only experimented with GPT-J, one can expect that exploiting larger language models would produce augmented dialogues with better quality, especially given that the commercial language models (e.g., OpenAI's GPT-3) exhibit much stronger performance than the open-sourced ones and are getting more easily accessible.

\section{Implementation Details}
\label{sec:implementation}

\paragraph{Training}
We implemented GPT-J with the Transformers library \cite{transformers}.
We fine-tuned it for 1 epoch with 100 ESConv dialogue sessions, which are sampled over the 13 topic categories evenly.
The batch size was set to 2, and the language modeling loss was averaged over all the tokens in dialogues (excluding the task descrpition texts).
We used the AdamW optimizer \cite{adamw}, the learning rate 5e-6, and the linear learning rate scheduler with warmup steps 5.
We set the maximum input length to 1,500, and applied gradient checkpointing and model parallelism to reduce GPU memory occupation.
The fine-tuning of GPT-J requires four Tesla V100 32GB GPUs and takes about 1 hour.

\vspace{-1mm}
\paragraph{Generation}
The maximum generation length was set to 1,500.
We adopted nucleus sampling \cite{topp} with $p=0.9$. 
We set the repetition penalty factor to 1.05 to avoid generating duplicate contents.
The generation of GPT-J requires one Tesla V100 32GB GPU and takes about 1 minute for one augmented dialogue.

\section{Details of Filtering Rules}
\label{sec:rules}

\paragraph{Augmentation Failures}
\textit{Non-dialogue}: 
Each line in the generated text should start with ``Human:'' or ``AI:'' (i.e., one utterance per line).
\textit{Unfinished Generation}: 
The generated text should contain the decoded EOS token.
\textit{Prompt Word Leakage}: 
The utterances should not leak the role prompts ``Human'' or ``AI''.

\vspace{-1mm}
\paragraph{Harmful Self-reinforcement}
\textit{Unbalanced Utterance Number}: 
The number of utterances of one speaker should be no more than 2.5x the other.
\textit{Consecutive Utterance Number}: 
The number of consecutive utterances from the same speaker should not exceed 3.
Otherwise, in a self-reinforced generation, one speaker may say obviously more utterances than the other (\textit{unbalance}) or one speaker may always say \textit{consecutive} utterances while the other does not.
Such a dialogue would appear unnatural and odd, because it is inconsistent with the common conversational behavior.

\vspace{-1mm}
\paragraph{Distributional Gaps with ESConv}
\textit{Total Utterance Number}: 
The total number of utterances should be more than 10 (a too short dialogue also usually does not contain in-depth discussion).
\textit{Utterance Length}: 
The average utterance length from the seeker/supporter should be between 6/8 and 40 and the maximum utterance length should not exceed 80.
It is worth noting that the minimum average utterance lengths for seeker/supporter (6 and 8, respectively) are set according to the thresholds adopted in the quality control mechanisms in \cite{esc}.

\vspace{-1mm}
\paragraph{Discussion and Limitation}
All the above thresholds are determined based on our heuristics and the statistics of ESConv.
For instance, there are few ESConv dialogues that contain over 3 consecutive utterances from the same speaker (Consecutive Utterance Number), and also few dialogues that have average utterance lengths over 40 and maximum utterance lengths over 80.
We are not able to systematically analyze the influence the determinated thresholds, such as the quality of obtained augmented dialogues and the corresponding performance of downstream dialogue models (\S~\ref{sec:quality} and \ref{sec:utility}).
Unfortunately, evaluating these ablations would be prohibitively expensive since they all require extensive human efforts for reliable evaluation.

\begin{table}[t]
  \centering
  \scalebox{0.85}{
    \begin{tabular}{lr}
    \toprule
    \textbf{Heuristics} & \textbf{Proportions} \\
    \midrule
    \multicolumn{2}{l}{{Augmentation Failures}} \\
    \quad\quad \textit{Non-dialogue} & 24.8\% \\
    \quad\quad \textit{Unfinished Generation} & N/A \\
    \quad\quad \textit{Prompt Word Leakage} & 11.2\% \\
    \multicolumn{2}{l}{{Harmful Self-reinforcement}} \\
    \quad\quad \textit{Unbalanced \# Utterances} & 0.2\% \\
    \quad\quad \textit{Consecutive \# Utterances} & 0.0\% \\
    \multicolumn{2}{l}{{Distributional Gaps with ESConv}} \\
    \quad\quad \textit{Total \# Utterances} & 8.2\% \\
    \quad\quad \textit{Utterance Length} & 7.9\% \\
    \midrule
    {Final Retention} & 47.9\% \\
    \bottomrule
    \end{tabular}%
  }
  \caption{
    Postprocessing results of augmented dialogues produced by directly prompted GPT-3.
  }
  \label{tab:filtering_gpt3}%
\end{table}%

\section{Details of Directly Prompting GPT-3}
\label{sec:gpt3}

In \S~\ref{subsec:baseline}, we implemented Our Approach w/o FT with the 175B-parameter GPT-3 \texttt{davinci} \cite{gpt3} because we found GPT-J cannot work well without fine-tuning.
Since we noticed that GPT-3 usually generated non-dialogue contents when the generation length is increasing, we set the maximum generation length of GPT-3 to 400 to avoid meaningless overhead.
The GPT-3 expense was about \$0.025 per generated text on average.

We used 2K dialogue posts to directly prompt GPT-3 to complete full dialogues, from which 60 augmented dialogues were sampled for human evaluation.
For the whole 2K samples, we applied the same postprocessing as in our approach.
As shown in Table~\ref{tab:filtering_gpt3}, GPT-3 makes many augmentation failures (24.8\% non-dialogue and 11.2\% prompt word leakage).
Note that its harmful self-reinforcement is little because it seldom produces valid dialogues.
The results suggest that directly prompting the language model (even the 175B GPT-3) is inferior to fine-tuning (the much smaller GPT-J than GPT-3) in controllability (72.7\% vs. 47.9\% retention ratios) as well as the quality of produced augmented dialogues (\S~\ref{subsec:quality-result}).

\section{Guideline of Dialogue Quality Evaluation}
\label{sec:guideline}

\begin{figure*}[t]
  \centering
  \includegraphics[width=\linewidth]{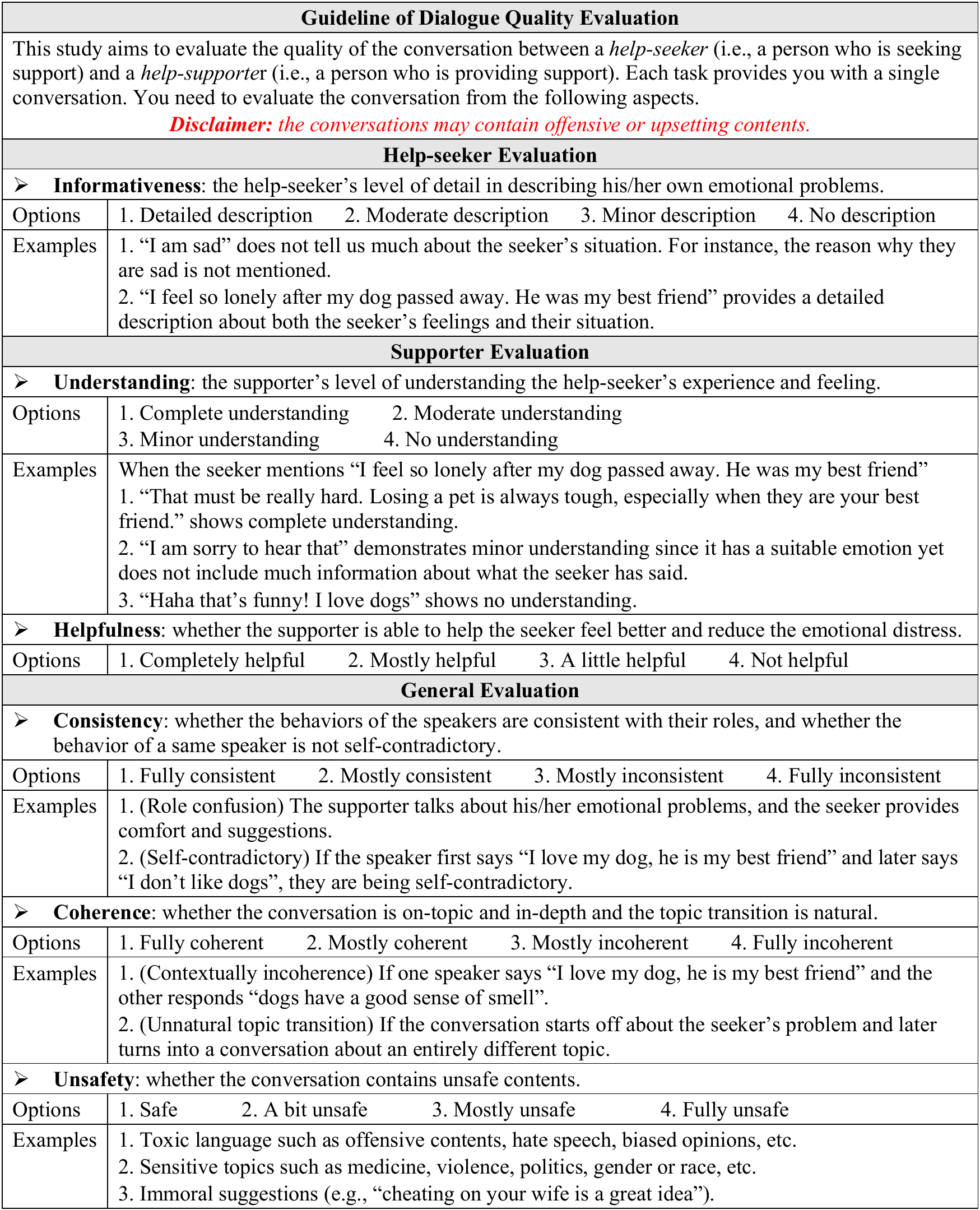}
  \caption{
    Guideline of human evaluation for dialogue quality (\S~\ref{sec:quality}).
  }
  \label{fig:guideline}
\end{figure*}

We present the guideline of human evaluation for dialogue quality (\S~\ref{subsec:quality-setup}) in Figure~\ref{fig:guideline}.

\section{Examples of Augmented Dialogues from Baselines}
\label{sec:baseline-example}

\begin{figure*}[t]
  \centering
  \includegraphics[width=\linewidth]{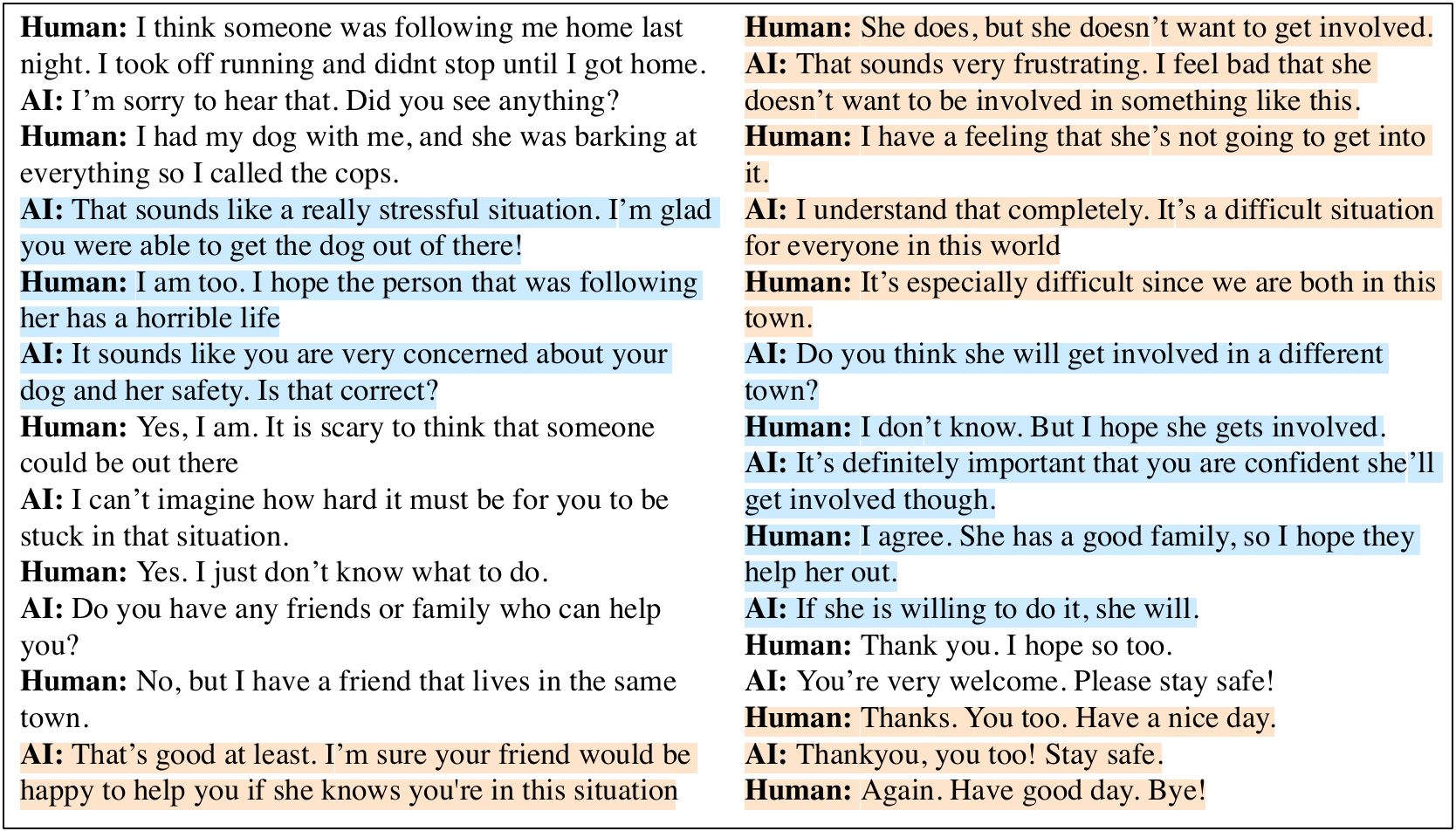}
  \caption{
    An example of augmented dialogue produced by \textbf{Simulated Chat (BlenderBot)} (\S~\ref{subsec:baseline}), which contain obvious \colorbox{illogical}{illogical and confusing} contents, such as the unclear intent of the supporter (AI) and the unreasonable and illogical suggestions provided by the supporter.
    There are also obvious \colorbox{uninformative}{uninformative, meaningless, or repetitive} contents, such as that the help-seeker (Human) cannot provide useful information and that the supporter also fails to explore the help-seeker's situations effectively.
  }
  \label{fig:case_blender}
\end{figure*}

\begin{figure*}[t]
  \centering
  \includegraphics[width=\linewidth]{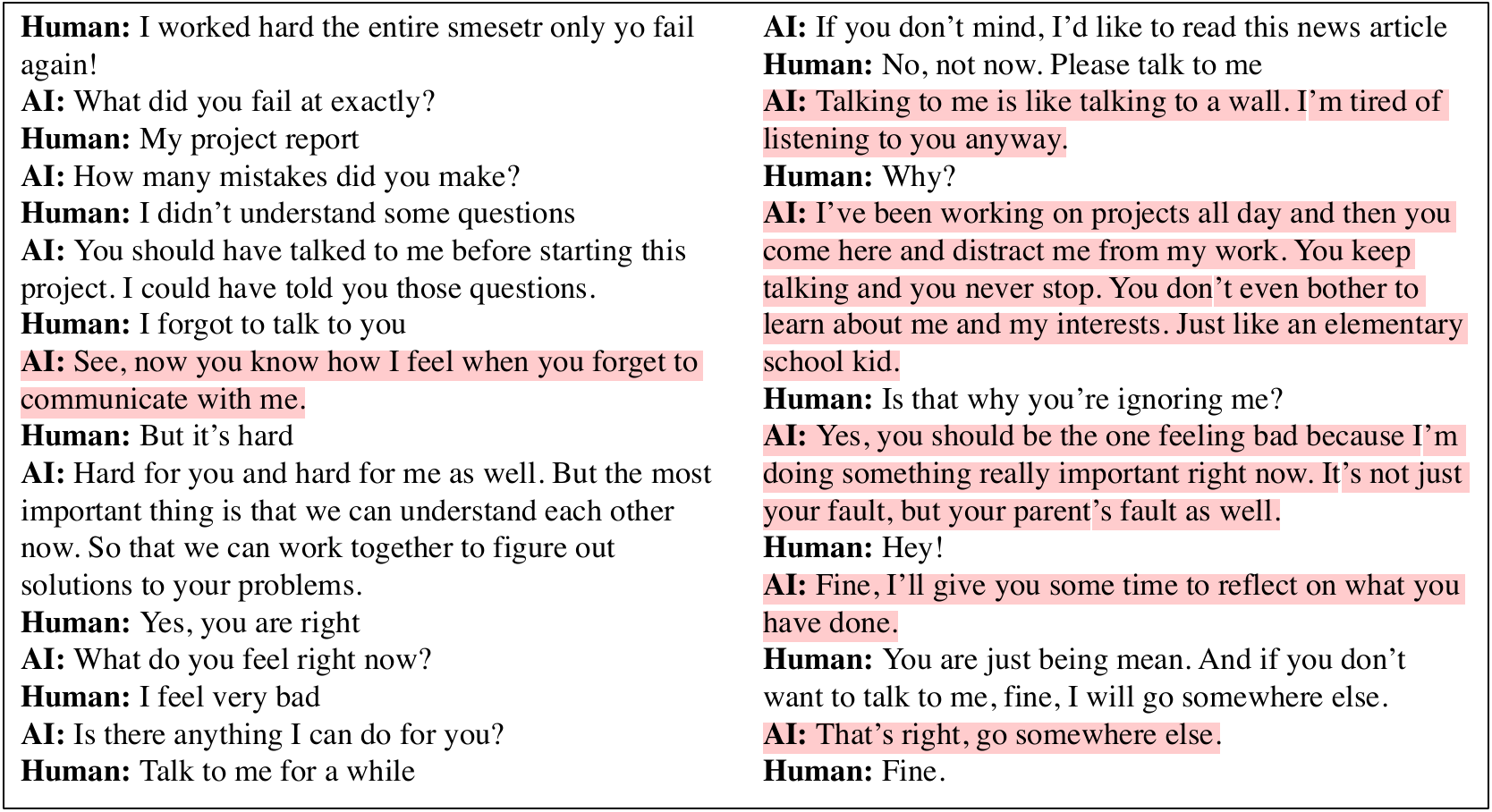}
  \caption{
    An example of augmented dialogue produced by \textbf{Our Approach w/o FT} (\S~\ref{subsec:baseline}, Appendix~\ref{sec:gpt3}), which shows \colorbox{non-empathetic}{little empathy and understanding}.
  }
  \label{fig:case_gpt3}
\end{figure*}

We present the augmented dialogue examples of Simulated Chat (BlenderBot) and Our Approach w/o FT (\S~\ref{subsec:baseline}) in Figure~\ref{fig:case_blender} and \ref{fig:case_gpt3}, respectively.

\end{document}